\documentclass{article}

\PassOptionsToPackage{numbers, compress}{natbib}
\usepackage[final]{neurips_2021}
\usepackage{caption}
\usepackage{subfigure}

\usepackage[utf8]{inputenc} 
\usepackage[T1]{fontenc}    
\usepackage{hyperref}       
\usepackage{url}            
\usepackage{booktabs}       
\usepackage{amsfonts}       
\usepackage{nicefrac}       
\usepackage{microtype}      
\usepackage{xcolor}         

\usepackage{wrapfig}
\usepackage{graphicx}
\usepackage[normalem]{ulem}
\useunder{\uline}{\ul}{}
\usepackage{amsmath}
\usepackage{amsthm}
\usepackage{float}

\usepackage{colortbl}
\newcommand{\red}{\textcolor{red}}
\newcommand{\blue}{\textcolor{blue}}

\usepackage{lipsum}                     
\usepackage{xargs}                      
\usepackage{tablefootnote}
\usepackage{amssymb}

\newtheorem{statement}{Statement}
\newtheorem{definition}{Definition}

\title{Overlapping Spaces for Compact \\ Graph Representations}

\author{%
  Kirill Shevkunov \\
    Yandex, MIPT \\
    Moscow, Russia \\
  \texttt{shevkunov.ks@phystech.edu} \\
  \And
   Liudmila Prokhorenkova \\ 
    Yandex Research, MIPT, HSE University \\
    Moscow, Russia \\
  \texttt{ostroumova-la@yandex.ru} \\
}

\begin{document}

\maketitle

\begin{abstract}
    Various non-trivial spaces are becoming popular for embedding structured data such as graphs, texts, or images. Following spherical and hyperbolic spaces, more general \emph{product spaces} have been proposed. However, searching for the best configuration of a product space is a resource-intensive procedure, which reduces the practical applicability of the idea. We generalize the concept of product space and introduce an \emph{overlapping space} that does not have the configuration search problem. The main idea is to allow subsets of coordinates to be shared between spaces of different types (Euclidean, hyperbolic, spherical). As a result, we often need fewer coordinates to store the objects. Additionally, we propose an optimization algorithm that automatically learns the optimal configuration. Our experiments confirm that overlapping spaces outperform the competitors in graph embedding tasks with different evaluation metrics. We also perform an empirical analysis in a realistic information retrieval setup, where we compare all spaces by incorporating them into DSSM. In this case, the proposed overlapping space consistently achieves nearly optimal results without any configuration tuning. This allows for reducing training time, which can be essential in large-scale applications.
\end{abstract}


\section{Introduction}\label{sec:introduction}

Building vector representations of various objects is one of the central tasks of machine learning. Word embeddings such as Glove~\cite{pennington2014glove} and Word2Vec~\cite{mikolov2013word2vec} are widely used in natural language processing; a similar Prod2Vec~\cite{grbovic2015commerce} approach is used in recommendation systems. There are many algorithms proposed for graph embeddings, e.g., Node2Vec~\cite{grover2016node2vec} and DeepWalk~\cite{perozzi2014deepwalk}. Recommendation systems often construct embeddings of a bipartite graph that describes interactions between users and items~\cite{yifan2009als}.

For a long time, embeddings were considered exclusively in $\mathbb{R}^n$. However, the hyperbolic space was shown to be more suitable for graph, word, and image representations due to the underlying hierarchical structure~\cite{khrulkov2020hyperbolic,nickel2017poincare,nickel2018learning,tifrea2018poincar}. Going beyond spaces of constant curvature, a recent study~\cite{gu2018learning} proposed \emph{product spaces}, which combine several copies of Euclidean, spherical, and hyperbolic spaces. While these spaces demonstrate promising results, the optimal signature (types of combined spaces and their dimensions) has to be chosen via brute force, which may not be acceptable in large-scale applications.

In this paper, we propose a more general metric space called \emph{overlapping space} (OS) together with an optimization algorithm that trains signature \emph{simultaneously} with embedding allowing us to avoid brute-forcing. The main idea is to allow coordinates to be shared between different spaces, which significantly reduces the number of coordinates needed.

Importantly, the proposed overlapping space can further be enhanced by adding non-metric approaches such as~\emph{weighted inner product}~\cite{kim2019representation} as additional similarity measures complementing metric ones. Thus, we obtain a flexible hybrid measure \emph{OS-Mixed} that is no longer a metric space. Our experiments show that in some cases, non-metric measures outperform metric ones. The proposed OS-Mixed has advantages of both worlds and thus achieves superior performance.


To validate the usefulness of the proposed overlapping space, we provide an extensive empirical evaluation for the task of graph embedding, where we consider both distortion-based (i.e., preserving distances) and ranking-based (i.e., preserving neighbors) objectives. In both cases, the proposed measure outperforms the competitors. We also compare the spaces in information retrieval and recommendation tasks, for which we apply them to train embeddings via DSSM~\cite{huang2013learning}. Our method works comparable to the best product spaces tested in these cases, while it does not require brute-forcing for the best signature. Thus, using the overlapping space may significantly reduce the training time, which can be crucial in large-scale applications.






\section{Background and related work}\label{sec:bacground}

\subsection{Embeddings and loss functions}\label{sec:loss_function}

For a graph $G = (V, E)$ an embedding is a mapping $f : V \rightarrow U$, where $U$ is a metric space equipped with a distance $d_U: U \times U \rightarrow \mathbb{R}_+$.\footnote{Note that any discrete metric space corresponds to a weighted graph, so graph terminology is not restrictive.} On the graph, one can consider a shortest path distance $d_G : V \times V \rightarrow \mathbb{R}_+$. In the graph reconstruction task, it is expected that a good embedding  preserves the original graph distances: $d_G(v, u) \approx d_U(f(v), f(u))$. The most commonly used evaluation metric is \emph{distortion}, which averages relative errors of distance reconstruction over all pairs of nodes:
\begin{equation}\label{distortion}
D_{avg} = \frac{2}{|V|(|V| - 1)}\sum\limits_{(v, u) \in V^2, v \neq u} \frac{|d_U(f(v), f(u)) - d_G(v, u)|}{d_G(v, u)}\,.
\end{equation}
While commonly used in graph reconstruction, distortion is a natural choice for many practical applications. 
For example, in recommendation tasks, one usually deals with a partially observed graph (some positive and negative element pairs), so a huge graph distance between two nodes in the observed part does not necessarily mean that the nodes are not connected by a short path in the full graph.
Also, often only the order of the nearest elements is essential while predicting distances to faraway objects is not critical.
In such cases, it is more reasonable to consider a local ranking metric, e.g., the mean average precision (mAP) that measures the relative closeness of the relevant (adjacent) nodes compared to the others:\footnote{For mAP, the relevance labels are assumed to be binary (unweighted graphs). If a graph is weighted, then we say that $N_v$ consists of the closest element to $v$ (or several closest elements if the distances to them are equal).}
\begin{equation}\label{map}
   \begin{aligned}
   &\mathrm{mAP} = \frac{1}{|V|}\sum\limits_{v \in V} \mathrm{AP}(v)= \frac{1}{V}\sum\limits_{v \in V}  \frac{1}{{\rm deg}(v)}\sum\limits_{u \in N_v} \frac{|N_v \cap R_{v}(u)|}{|R_{v}(u)|}, \\
    &R_{v}(u) = \{w \in V | d_U\big{(}f(v), f(w)\big{)} \leq d_U\big{(}f(v), f(u)\big{)}\},\, N_v = \{w \in V | (v, w) \in E\}.
    \end{aligned}
\end{equation}

Mean average precision cannot be directly optimized since it is not differentiable. In our experiments, we use the following probabilistic loss function as a proxy:\footnote{See Table~\ref{sdplc-table} in Appendix for the comparison of other ways of converting distance to probability.}
\begin{equation}\label{proxyloss}
    \begin{aligned}
    L_{proxy} = - \sum\limits_{(v, u) \in E} \log \mathrm{P}((v, u) \in E)
    = - \sum\limits_{(v, u) \in E} \log \frac{\exp(-d_U(f(v), f(u)))}{\sum\limits_{w \in V}\exp(-d_U(f(v), f(w)))}\,.
    \end{aligned}
\end{equation}
Note that when substituting $d_U(x, y) = c - f(x)^Tf(y)$ (assuming that $f(x) \in \mathbb{R}^n$, so the dot product is defined), $L_{proxy}$ becomes the standard word2vec loss function.

\subsection{Spaces, distances, and similarities}\label{sec:distances}

In the previous section, we assumed that $d_U: U \times U \rightarrow \mathbb{R}_+$ is an arbitrary distance. In this section, we discuss particular choices often considered in the literature.

For many years, Euclidean space was the primary choice for structured data embeddings~\cite{goyal2018graph}. For two points $x,y \in \mathbb{R}^d$, Euclidean distance is defined as $d_E(x, y) = \left(\sum\limits_{i=1}^d (x_i - y_i)^2\right)^{1/2}$.

Spherical spaces were also found to be suitable for some applications~\cite{liu2017sphereface,qian2004cad,wilson2014spherical}. Indeed, in practice, vector representations are often normalized, so cosine similarity between vectors is a natural way to measure their similarity. This naturally corresponds to a spherical space $S_d = \{x \in \mathbb{R}^{d+1}: \|x\|_2^2 = 1\}$ equipped with a distance $d_S(x,y) = \arccos (x^T y)$.

In recent years, hyperbolic spaces also started to gain popularity.
Hyperbolic embeddings have shown their superiority over Euclidean ones in a number of tasks, such as graph reconstruction and word embedding~\cite{nickel2017poincare,nickel2018learning,sala2018representation,tifrea2018poincar}.
To represent the points, early approaches used the Poincar\'e model of the hyperbolic space~\cite{nickel2017poincare}, but later it has been shown that the hyperboloid (Lorentz) model may lead to more stable results~\cite{nickel2018learning}. 
In this work, we also adopt the hyperboloid model $H_d = \{x \in \mathbb{R}^{d+1} | \langle x, x\rangle_h = 1, x_1 > 0\}$ equipped with a distance $d_{H} = \mathrm{arccosh}(\langle x, y \rangle_h)$, where $\langle x, y \rangle_h := x_1 y_1 - \sum\limits_{i=2}^{d+1} x_i y_i$.

Going even further, a recent paper~\cite{gu2018learning} proposed more complex \emph{product spaces} that combine several copies of Euclidean, spherical, and hyperbolic spaces. 
Namely, the overall dimension $d$ is split into $k$ parts (smaller dimensions): $d = \sum\limits_{i=1}^{k}d_i$, $d_i > 0$. Each part is associated with a space $D_i \in \{E_{d_i}, S_{d_i}, H_{d_i}\}$ and scale coefficient $w_i \in \mathbb{R}_+$. Varying scale coefficients corresponds to changing curvature of hyperbolic and spherical spaces, while in Euclidean space this coefficient is not used ($w_i = 1$). Then, the distance in the product space $D_1 \times \ldots \times D_k$ is defined as:
$$
d_{P}(x, y)
= \sqrt{\sum\limits_{i=1}^{k} w_i \, d_{D_i}(x[t_{i-1}+1:t_{i}], y[t_{i-1}+1:t_{i}])^2}\,,
$$
where $t_0 = 0$, $t_i = t_{i-1} + d_i$, and $x[s:e]$ is a subvector $(x_{s}, \ldots, x_e) \in  \mathbb{R}^{e-s+1}$. If $k = 1$, we get a standard Euclidean, spherical, or hyperbolic space.

In~\cite{gu2018learning}, it is proposed to learn an embedding and scale coefficients $w_i$ simultaneously. However, choosing the optimal signature (how to split $d$ into $d_i$ and which types of spaces to choose) is challenging. A heuristics proposed in~\cite{gu2018learning} allows to guess types of spaces if $d_i$'s are given. If $d_1 = d_2 = 5$, this heuristics agrees well with the experiments on three considered datasets. The generalizability of this idea to other datasets and configurations is unclear. In addition, it cannot be applied if a dataset is partially observed and graph distances cannot be computed (e.g., when there are several known positive and negative pairs). Hence, in practice, it is more reliable to choose a signature via the brute-force, which can be infeasible on large datasets.

Another way to measure objects' similarity, which is rarely compared with metric methods but is frequently used in practical applications, is via the dot product of vectors $x^Ty$ or its weighted version $x^TWy$ with a diagonal matrix $W$, which is also known as a weighted inner product~\cite{kim2019representation}. Such measures cannot be converted to a metric distance via a monotone transformation. However, they can be used to predict similarity or dissimilarity between objects, which is often sufficient in practice, especially when ranking metrics are used.

In this paper, we stress that when comparing different methods, both metric and non-metric variants should be used because different methods are better for different tasks. In particular, the dot product similarity allows one to easily differentiate between more popular and less popular items (the vector norm can be considered a measure of popularity). This feature is also attributed to hyperbolic spaces, where more popular items are placed closer to the origin.

\subsection{Optimization} \label{ss:psopt}

Gradient optimization in Euclidean space is straightforward, while for spherical or hyperbolic embeddings, we also have to control that the points belong to a surface. In previous works, Riemann-SGD was used to solve this problem~\cite{bonnabel2013StochasticGD}. In short, it projects Euclidean gradients on the tangent space at a point and then uses a so-called exponential map to move the point along the surface according to the gradient projection. For product spaces, a generalization of the exponential map has been proposed~\cite{ficken1939psdiffgeom, riemannian2016cv}.

In~\cite{wilson2018gradient}, the authors compare RSGD with the retraction technique, where points are moved along the gradients in the ambient space and are projected onto the surface after each update. From their experiment, the retraction technique requires from 2\% to 46\% more iterations, depending on the learning rate. However, the exponential update step takes longer. Hence, the advantage of RSGD in terms of computation time depends on the specific implementation.

\section{Overlapping spaces}\label{sec:overlapping}

\subsection{Overlapping metric spaces}

In this section, we propose a new concept of \emph{overlapping spaces}.\footnote{We refer to Appendix~\ref{sec:os-with-pictures} for additional simple illustrations of the proposed idea.}\label{sec:overlapping-itself} This approach generalizes product spaces and allows us to make the signature (types and dimensions of combined spaces) trainable. Our main idea is to divide the embedding vector into several \emph{overlapping} (unlike product spaces) segments, each segment corresponding to its own space. Then, instead of discrete signature brute-forcing, we optimize the weights of the signature elements. 

Importantly, we allow the same coordinates of an embedding vector to be included in distances computations for spaces of different geometry. For this purpose, we first need to map a vector $x \in \mathbb{R}^{d}$ (for any $d\ge 1$) to a point in Euclidean, hyperbolic, and spherical space. Let us denote this mapping by $M$. Obviously, for Euclidean space, we may take $M_E(x) = x$. We may use the vector normalization for the spherical spaces, and for $H_d$ we use a projection from a hyperplane to a hyperboloid:
\begin{equation}\label{eq:mapping}
    M_S(x) = \frac{x}{|x|} \in S_{d-1}, \,\, M_H(x) = \left(\sqrt{1 + \sum\limits_{i=2}^d x_i^2}, x_1, \ldots, x_d\right) \in H_d. 
\end{equation}
Note that for such parametrization a $d$-dimensional vector $x$ is mapped into Euclidean and hyperbolic spaces of dimension $d$ and into a spherical space of dimension $d-1$. Hence, in standard implementations of product spaces, a sphere $S_d$ is parametrized by a $(d+1)$-dimensional vector~\cite{gu2018learning}.
However, this requires more coordinates to be stored for each spherical space. Hence, to make a fair comparison of all spaces, we use the hyperspherical coordinates for $S_d$:
\begin{equation}\label{eq:hypersperical}
    \setlength{\tabcolsep}{1pt}
    \hat{M}_S(x) = \begin{pmatrix}
        \begin{array}{l}
            \cos x_1 \cdot \cos x_2 \cdot  \ldots  \cdot \cos x_{d-1} \cdot \cos x_d \\
            \cos x_1 \cdot \cos x_2 \cdot  \ldots  \cdot \cos x_{d-1} \cdot \sin x_d \\
            \cos x_1 \cdot \cos x_2 \cdot  \ldots  \cdot \sin x_{d-1}   \\
            \,\,\,\, \ldots  \\
            
            \sin x_1  \\
        \end{array}
    \end{pmatrix} \in S_d.
\end{equation}
Now we are ready to define an overlapping space. Consider two vectors $x, y \in \mathbb{R}^d$. Let $p_1, \ldots, p_k$ denote some subsets of coordinates, i.e., $p_i \subset \{1, \ldots, d\}$. We assume that together these subsets cover all coordinates, i.e., $\cup_{i=1}^k p_i = \{1, \ldots, d\}$. By $x[p_i]$ we denote a subvector of $x$ induced by $p_i$. Let $D_i \in \{E, S, H\}$. We define $d_i(x,y) := d_{D_i} \big{(}M_{D_i}(x[p_i]),M_{D_i}(y[p_i]) \big{)}$ and aggregate these distances with arbitrary positive weights $w_1 \ldots w_k \in \mathbb{R}_+$:
\begin{equation}\label{eq:overlaying_distance}
\begin{aligned}
    d_{O}^{l1}(x, y) &= \sum\limits_{i=1}^k w_i d_i(x, y) \,, \\
    d_{O}^{l2}(x, y) &= \left(\sum\limits_{i=1}^k w_i d_i^2(x, y)\right)^{1/2}.
\end{aligned}
\end{equation}

\begin{definition}
$O_d = \{x \in \mathbb{R}^{d}\}$ equipped with a distance $d_{O}^{l1}$ or $d_{O}^{l2}$ defined in~\eqref{eq:overlaying_distance} is called an \emph{overlapping space}. This space is defined by $p_i$, $D_i$, and $w_i$.
\end{definition}

Note that it is sufficient to assume that spherical and hyperbolic spaces have curvatures $1$ and $-1$, respectively, since changing the curvature is equivalent to changing scale, which is captured by $w_i$. The following statement follows from the definition above and the fact that $d_E$, $d_S$, and $d_H$ are distances (see Appendix~\ref{sec:os-is-metric-proof} for the proof).

\begin{statement}\label{st:metric}
If $\cup_{i=1}^k p_i = \{1, \ldots, d\}$ and $w_1 \ldots w_k \in \mathbb{R}_+$, then $d_{O}^{l1}$, $d_{O}^{l2}$ are distances on $\mathbb{R}^d \times \mathbb{R}^d$, i.e., they satisfy the metric axioms.
\end{statement}

It is easy to see that overlapping spaces generalize product spaces. Indeed, if we assume $p_i \cap p_j = \emptyset$ for all $i \neq j$, then an overlapping space reduces to a product space. However, the fact that we allow  $p_i \cap p_j \neq \emptyset$ gives us a significantly larger expressive power for the same dimension $d$.

\subsection{Generalization with WIPS: OS-Mixed measure}\label{os-mixed}

It is known that for ranking loss functions, methods based on weighted and standard dot products can have good performance~\cite{kim2019representation}. Let us note that such similarity measures cannot be converted to a distance via a monotone transformation. In our experiments, we notice that in some cases, such non-metric methods can successfully be used for graph embeddings even with the distortion loss, when the model approximates metric distances.

To close the gap between metric and non-metric methods, we propose a generalization of the overlapping spaces that also includes the weighted inner product similarity (WIPS).

First, let us motivate our choice of WIPS. One may extend the list of base distance functions $\{d_E, d_S, d_H\}$ with a dissimilarity measure $d_{\text{dot}} = c - x^Ty$ that is not a metric distance. By using such `distance' for all possible subsets $p_i \in 2^{\{1\ldots d\}}$ and applying $l1$-aggregation~\eqref{eq:overlaying_distance}, we get WIPS measure $d_W = \tilde c -  \sum\limits_{i=1}^{d} \tilde w_i x_i y_i$, where $\tilde c$ and $\tilde w_i$ are trainable values.

Thus, instead of using an extended set of base distances $\{d_E, d_S, d_H, d_{\text{dot}}\}$ together, as shown in equation~\eqref{eq:overlaying_distance}, we suggest to simply use $d_{OS-Mixed}(x, y) = d_{O}(x, y) + d_{W}(x, y)$ where $d_{O}$ is a metric overlapping space. We show that this design gives excellent results for both distortion and ranking objectives in the graph reconstruction task. We further refer to this approach as \emph{OS-Mixed}.




\section{Optimization in overlapping spaces}

\subsection{Binary tree signature}

Overlapping spaces defined in Section~\ref{sec:overlapping} are flexible and allow capturing various geometries. However, similarly to product spaces, they need a signature ($p_i$ and $D_i$) to be chosen in advance. This section describes a particular signature, which is still flexible, does not suffer from the brute-force problem, and shows promising empirical results on the datasets we experimented with.

\begin{figure}
  \begin{center}
    \includegraphics[width=0.9\textwidth]{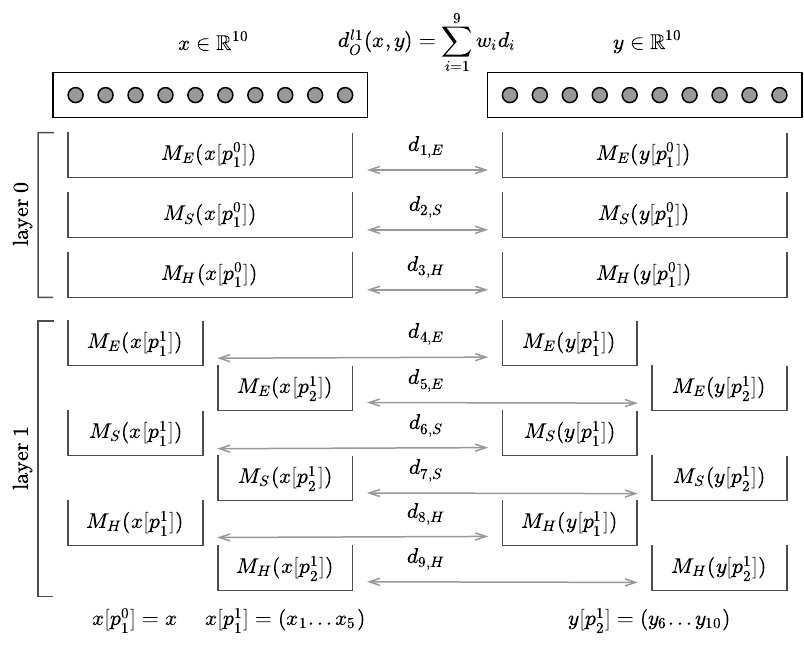}
  \end{center}
  \caption{Overlapping space with $d=10$, $t=1$, and $l1$ (sum) aggregation}\label{fig:ttl1}
\end{figure}

Let $t \ge 0$ denote the depth (complexity) of the signature for a $d$-dimensional embedding. 
Each layer $l$, $0 \le l \le t$, of the signature  consists of $2^l$ subsets of coordinates: $p_i^l = \left\{\left[d(i-1)/2^{l}\right]+1, \ldots, \left[di/2^{l}\right] \right\}\,,\,\,\,\, 1 \le i \le 2^l\,$. Each $p_i^l$ is associated with Euclidean, spherical, and hyperbolic spaces simultaneously. The corresponding weights are denoted by 
$w_i^{l,E}, w_i^{l,S}, w_i^{l,H}$.
Then, the distance is computed according to~\eqref{eq:overlaying_distance}. 
See Figure~\ref{fig:ttl1} for an illustration of the procedure for $d = 10$ and $t = 1$.

Informally, assume that we have two vectors $x,y \in \mathbb{R}^d$. To compute the distance between them in the proposed overlapping space, we first compute Euclidean, spherical, and hyperbolic distances between $x$ and $y$. Then, we split the coordinates into two halves, and for each half, we also compute all three distances, and so on. Finally, all the obtained distances are averaged with the weight coefficient according to~\eqref{eq:overlaying_distance}.
Note that we have $3(2^{t+1} - 1)$ different weights in our structure in general, but with $l2$-aggregation this value may be reduced to $2(2^{t+1} - 1) + 2^t$ since for the Euclidean space, the distances between subvectors at the upper layers can be split into terms corresponding to smaller subvectors, so we essentially need only the last layer with $2^t$ terms.

Recall that in a product space, the weights correspond to curvatures of the combined hyperbolic and spherical spaces. In our case, they also play another important role: the weights allow us to balance between different spaces. Indeed, for each subset of coordinates, we simultaneously compute the distance between the points assuming each of the combined spaces. Varying the weights, we can increase or decrease the contribution of a particular space to the distance. As a result, our signature allows us to learn the optimal combination, which does not have to be a product space since all weights can be non-zero. Interestingly, when we analyzed the optimized weights, we observed that often some of them are close to zero. Thus, unnecessary weights can be detected and removed after the training. See Appendix~\ref{sec:os-weights} for more details.

Note that the procedure described in this section naturally extends to OS-Mixed by adding the corresponding `distance' to Euclidean, hyperbolic, and spherical as described in Section~\ref{os-mixed}.

\subsection{Optimization}

In this section, we describe how we embed into the overlapping space.
Although Riemann-SGD (see Section \ref{ss:psopt}) is a good solution from the theoretical point of view, in practice, due to errors in storing and processing real numbers, it may cause some problems. Indeed, a point that we assume to lie on a surface (sphere or hyperboloid) does not numerically lie on it usually. Due to the accumulation of numerical errors, with each iteration of RSGD, a point may move away from the surface. Therefore, in practice, after each step, all embeddings are explicitly projected onto the surface, which may slow down the algorithm. Moreover, RSGD is not applicable if one needs to process the output of a neural network, which cannot be required to belong to a given surface (e.g., to satisfy $\langle x, x\rangle_h = 1 \Leftrightarrow x \in H_{d}$). As a result, before finding the hyperbolic distance between two outputs of a neural network in the Siamese~\cite{bromley1994signature} setup, one first needs to map them to a hyperboloid.

Instead of RSGD, we store the embedding vectors in Euclidean space and calculate distances between them using the mappings~\eqref{eq:mapping} to the corresponding surfaces. Thus, we can evaluate the distances between the outputs of neural networks and also use conventional optimizers. To optimize embeddings, we first map Euclidean vectors into the corresponding spaces, calculate distances and loss function, and then backpropagate through the projection functions. To improve the convergence, we use Adam~\cite{kingma2014adam} instead of the standard SGD. Similar technique using SGD with momentum is used in~\cite{Law2019LorentzianDL}. Applying this to product spaces, we achieve the results similar to the original paper~\cite{gu2018learning} (see Table~\ref{original-table} 
in Appendix), where RSGD was used with the learning rate brute-forcing, custom learning rate for curvature coefficients, and other tricks.

\section{Experiments}\label{sec:experiments}

\subsection{Compared spaces}

In this section, we provide a thorough analysis to compare all metric spaces discussed in the paper, including product spaces with all signatures from~\cite{gu2018learning} and the proposed overlapping space. For the non-metric dissimilarity functions, we consider $d(x, y) = c - x^Ty$,  $d(x, y) = c - \sum w_i x_i y_i$ (WIPS), $d(x, y) = c\exp(- x^Ty)$ with trainable parameters $c, w_i \in \mathbb{R}$, and the proposed OS-Mixed measure. We add non-metric measures even to the distortion setup to see whether they are able to approximate graph distances. Similarly to~\cite{gu2018learning}, we fix the dimension $d=10$. However, for a fair comparison, we fix the number of \emph{stored values} for each embedding and use the hypersperical parametrization~\eqref{eq:hypersperical} instead of storing $d+1$ coordinates.\footnote{In Appendix~\ref{sec:sphere-param}, we evaluate spherical spaces without this modification to compare with~\cite{gu2018learning}.} The training details are given in Appendix~\ref{sec:setup}. The code of our experiments is available.\footnote{\href{https://github.com/shevkunov/overlapping-spaces-for-compact-graph-representations}{https://github.com/shevkunov/overlapping-spaces-for-compact-graph-representations}}

\subsection{Graph reconstruction}
\label{graph_rec}

\begin{table}
    \caption{Datasets for graph reconstruction}
    \label{dataset-table}
    \centering
    \begin{small}
    \begin{tabular}{l|cccccc}
        \toprule
        \textbf{}  & \textbf{USCA312} & \textbf{CS PhDs} & \textbf{Power} & \textbf{Facebook} & \textbf{WLA6} & \textbf{EuCore}\\
        \midrule
        Nodes
            & 312 & 1025 & 4941 & 4039 & 3227 & 986 \\
        Edges
            & 48516 (weighted) & 1043 & 6594 & 88234 & 3604 & 16687 \\
        \bottomrule
    \end{tabular}
    \end{small}
\end{table}

\paragraph{Graph datasets}
We use the following graph datasets: the USCA312 dataset of distances between North American cities \cite{burkardtXXXXcity} (weighted complete graph), a graph of computer science Ph.D. advisor-advisee relationships~\cite{bonacich2008phd}, a power grid distribution network with backbone structure~\cite{watts2011power}, a dense social network from Facebook~\cite{leskovec2012social}, and EuCore dataset generated using email data from a large European research institution~\cite{leskovec2007graph}. We also collected a new dataset by launching the breadth-first search on the Wikipedia category graph, starting from the ``Linear Algebra'' category with search depth limited to 6. Further, we refer to this dataset as WLA6; more details are given in Appendix~\ref{sec:WLA-dataset}.
This graph is very close to being a tree, although it has some cycles. We expect the hyperbolic space to give a significant profit for this graph, and we observe that product spaces give almost no additional advantage. The purpose of using this additional dataset is to evaluate overlapping spaces on a dataset where product spaces do not provide quality gains.
Table~\ref{dataset-table} lists the properties of all considered datasets.


\begin{table}
    \caption{Distortion graph reconstruction, top results are highlighted, top metric results are underlined}
    \label{ditortion-table}
    \centering
    \begin{small}
    \begin{tabular}{l|cccccc}
        \toprule
        \textbf{Signature}  & \textbf{USCA312} & \textbf{CS PhDs} & \textbf{Power} & \textbf{Facebook} & \textbf{WLA6} & \textbf{EuCore} \\
        \midrule
        $E_{10}$
            & \textbf{\underline{\blue{0.00318}}} & 0.0475 & 0.0408 & 0.0487 & 0.0530 & 0.1242 \\
        $H_{10}$
            & 0.01104 & 0.0443 & 0.0348 & 0.0483 & \underline{0.0279} & 0.1144 \\
        $S_{10}$
            & 0.01065 & 0.0519 & 0.0453 & 0.0561 & 0.0608 & 0.1260 \\
        $H_5^2 \equiv H_5\times H_5 $
            & 0.00573 & 0.0345 &  0.0255 & 0.0372 & \underline{0.0279} & \underline{0.1106} \\
        $S_5^2 \equiv S_5 \times S_5$
           & 0.00700 & 0.0501 & 0.0438 & 0.0552 & 0.0584 & 0.1251 \\
        $H_5 \times S_5$
            & 0.00541 & \underline{0.0341} & \underline{0.0254} & \underline{0.0346} & 0.0310 & 0.1195 \\
        $H_2^5$
            & 0.00592 & 0.0344 & 0.0273 & 0.0439 & 0.0356 & 0.1163 \\
        $S_2^5$
             & 0.00604 & 0.0464 & 0.0416 & 0.0512 & 0.0543 & 0.1244 \\
        $H_2^2 \times E_2 \times S_2^2$
            & 0.00537 & 0.0344 & 0.0302 & 0.0406 & 0.0437 & 0.1193 \\
        $O_{l1}, t=0$
            & \textbf{\underline{0.00324}} & 0.0368 & 0.0281 & 0.0458 & 0.0286 & 0.1141 \\
        $O_{l1}, t=1$
            & \underline{0.00325} & \textbf{\underline{0.0300}} & \textbf{\underline{\red{0.0231}}} & \underline{0.0371} & \underline{0.0272} & \underline{0.1117} \\
        $O_{l2}, t=1$
            & 0.00530 & \underline{0.0328} & \textbf{\underline{0.0246}} & \underline{0.0324} & \underline{0.0278} & \underline{0.1127} \\
        \midrule
        $c-\text{dot}$
            & 0.04005 & 0.0412 & 0.0461 & 0.0236 & 0.0296 & 0.1085 \\
        $c-\text{wips}$
            & 0.06468 & 0.0358 & 0.0442 & \textbf{\red{0.0161}} & \textbf{0.0238} & \textbf{\red{0.1016}} \\
        $c e^{-\text{dot}}$
            & 0.08142 & 0.0424 & 0.0505 & 0.0192 & 0.0270 & 0.1048  \\
        $O_{mix-l1}, t=1$
            & \textbf{\red{0.00277}} & \textbf{\blue{0.0243}} & \textbf{\blue{0.0235}} & \textbf{0.0172} & \textbf{\red{0.0187}} & \textbf{\blue{0.1026}} \\
        $O_{mix-l2}, t=1$
            & 0.00464 & \textbf{\red{0.0220}} & 0.0258 & \textbf{\blue{0.0163}} & \textbf{\blue{0.0198}} & \textbf{0.1028} \\
        \bottomrule
    \end{tabular}
    \end{small}
\end{table} 

\paragraph{Distortion loss}
We start with the standard graph reconstruction task with distortion loss~\eqref{distortion}. The goal is to embed all nodes of a given graph into a $d$-dimensional space approximating the pairwise graph distances between them. In this setup, all models are trained to minimize distortion~\eqref{distortion}, the results are shown in Table~\ref{ditortion-table}. It can be seen that the overlapping spaces outperform other metric spaces, and the best overlapping space (among considered) is the one with $l1$ aggregation and complexity $t=1$.\footnote{In Appendix~\ref{sec:os-more-t} we also analyze $t > 1$ and show that performance improves as $t$ increases.} Interestingly, the performance of such overlapping space is often better than the \emph{best} considered product space. 

We note that standard non-metric distance functions show highly unstable results for this task: for the USCA312 dataset, the obtained distortion is orders of magnitude worse than the best one. However, on some datasets (Facebook and WLA6), the performance is quite good, and for Facebook, these simple non-metric similarities have much better performance than all metric solutions. Thus, we conclude that such functions are worth trying for graph reconstruction with the distortion loss, but their performance is unstable. In contrast, the overlapping spaces show good and stable results on all datasets, and the proposed OS-Mixed modification (see Section~\ref{os-mixed}) outperforms all other approaches.

\paragraph{Ranking loss}
As discussed in Section~\ref{sec:loss_function}, in many practical applications, only the order of the nearest neighbors matters. In this case, it is more reasonable to use mAP~\eqref{map}. In previous work~\cite{gu2018learning}, mAP was also reported, but the models were trained to minimize distortion. In our experiments, we observe that distortion optimization weakly correlates with mAP optimization. Hence, we minimize the proxy-loss defined in equation~\eqref{proxyloss}. The results are shown in Table~\ref{map-table}, and the obtained values for mAP are indeed much better than the ones obtained with distortion optimization~\cite{gu2018learning}, i.e., it is essential to use an appropriate loss function. According to Table~\ref{map-table}, among the metric spaces, the best results are achieved with the overlapping spaces (especially for $l2$-aggregation with $t=1$). Importantly, in contrast to distortion loss, ranking based on the dot-product outperforms all metric spaces. However, using the OS-Mixed measure allows us to improve these results even further.

\begin{table}

    \caption{mAP graph reconstruction, top results are highlighted, top metric results are underlined}
    \label{map-table}
    \centering
    \begin{small}
    \begin{tabular}{l|cccccc}
        \toprule
        \textbf{Signature} & \textbf{USCA312} & \textbf{CS PhDs} & \textbf{Power}  & \textbf{Facebook} & \textbf{WLA6}  & \textbf{EuCore} \\
        \midrule
        $E_{10}$
            & 0.9290 & 0.9487 & 0.9380  & 0.7876 & 0.7199 & 0.6108 \\
        $H_{10}$
            & 0.9173 & 0.9399 & 0.9385 & 0.7997 & 0.9617 & 0.6670 \\
        $S_{10}$
            & 0.9183 & 0.9519 & 0.9445 & 0.7768 & 0.7289 & 0.6037 \\
        $H_5^2$
            & 0.9247 & 0.9481 & 0.9415 & 0.8084 & \underline{0.9682} & \underline{0.6783} \\
        $S_5^2$
            & 0.9316 & 0.9600 & 0.9482 & 0.7790 & 0.7307 & 0.6116 \\
        $H_5 \times S_5$
            & 0.9397 & 0.9538 & 0.9505 & 0.7947 & \underline{0.9751} & \underline{0.6847} \\
        $H_2^5$
            & 0.9364 & 0.9671 & 0.9508 & 0.7979 & 0.8597 & 0.6611 \\
        $S_2^5$
            & 0.9439 & 0.9656 & 0.9511 & 0.7800 & 0.7358 & 0.6169 \\
        $H_2^2 \times E_2 \times S_2^2$
            & 0.9519 & 0.9638 & 0.9507 & 0.7873  & 0.7794 & 0.6492 \\
        $O_{l1}, t=0$
            & \textbf{\underline{\blue{0.9538}}} & \underline{0.9879} & \underline{0.9728} & \underline{0.8093} & 0.6759 & 0.6580 \\
        $O_{l1}, t=1$
            & \textbf{\underline{0.9522}} & \textbf{\underline{0.9904}} & \underline{0.9762} & \underline{0.8185} & 0.9598 & 0.6691 \\
        $O_{l2}, t=1$
            & \textbf{\underline{0.9522}} & \textbf{\underline{\blue{0.9938}}} & \underline{0.9907} & \underline{0.8326} & \underline{0.9694} & \underline{0.7078} \\
        \midrule
        $c - \text{dot}$
            & \textbf{\red{1}} & \textbf{\red{1}} & \textbf{0.9983} & \textbf{0.8745} & \textbf{0.9990} & 0.7409 \\
        $c - \text{wips}$
            & \textbf{\red{1}} & \textbf{\red{1}} & \textbf{\red{1}} & 0.8704 & \textbf{\red{1}} & \textbf{0.7742} \\
        $O_{mix-l1}, t=1$
            & \textbf{\red{1}} & \textbf{\red{1}} & \textbf{\blue{0.9994}} & \textbf{\blue{0.8806}} & \textbf{\blue{0.9997}} & \textbf{\blue{0.7860}} \\
        $O_{mix-l2}, t=1$
            & \textbf{\red{1}} & \textbf{\red{1}} & \textbf{\red{1}} & \textbf{\red{0.9021}} & \textbf{\red{1}} & \textbf{\red{0.8405}} \\
        \bottomrule
    \end{tabular}
    \end{small}
\end{table}

\subsection{Information retrieval}
\label{dssm_exp}

\begin{table}
    \caption{DSSM embeddings, top three results are highlighted}
    \label{dssm-table}
    \vspace{4pt}
    \centering
    \begin{small}
        \begin{tabular}{llll}
        \toprule
        \textbf{Signature} & \textbf{Test mAP} \\
        \midrule
        $E_{10}$ & 0.4459 \\
        $H_{10}$ & 0.4047 \\
        $S_{10}$ & 0.4364 \\  
        $H_5^2$ & 0.4492 \\
        $S_5^2$ & \textbf{\blue{0.4573}} \\
        $H_5 \times S_5$ & 0.3295 \\
        $H_2^5$ & 0.3681 \\
        $S_2^5$ & \textbf{\red{0.4616}} \\
        $H_2^2 \times E_2 \times S_2^2$ & 0.3526 \\
        $c-\text{dot}$ & 0.4194 \\
        $O_{l1}, t=0$ & \textbf{0.4562} \\
        $O_{l1}, t=1$ & 0.4498 \\
        $O_{l2}, t=1$ & 0.4456 \\
        $O_{mix-l1}, t=1$ & 0.4447 \\
        $O_{mix-l2}, t=1$ & 0.4483 \\ 
        \bottomrule
    \end{tabular}
    \qquad\qquad
    \begin{tabular}{lll}
        \toprule
        \textbf{Signature} & \textbf{Test mAP} \\ 
        \midrule
        $E_{256}$
            & \textbf{\blue{0.717}} \\ 
        $H_{256}$
            & 0.412~\tablefootnote{The gap between $E_{256}$ and $H_{256}$ may seem suspicious, but in Table 5 of~\cite{gu2018learning} a similar pattern is observed.} \\ 
        $S_{255}$
            & 0.588 \\ 
        $H_{128}^2$
            & 0.547 \\ 
        $S_{127}^2$
            & 0.662 \\ 
        $H_{128} \times S_{127}$
            & 0.501 \\ 
        $H_{51}^4 \times H_{52}$
            & 0.621 \\ 
        $S_{50}^4 \times S_{51}$
            & \textbf{0.701} \\ 
        $c-\text{dot}$
            & \textbf{\red{0.738}} \\ 
        $O_{l1}, t=0$
            & 0.677 \\ 
        $O_{l1}, t=1$
            & 0.662 \\ 
        $O_{mix-l1}, t=1$
            & 0.663  \\
        $O_{mix-l2}, t=1$
            & 0.655  \\
            
        \bottomrule
    \end{tabular}
    \end{small}
\end{table}

From a practical perspective, it is also important to analyze whether an embedding can generalize to unseen examples. For instance, an embedding can be made via a neural network based on objects' characteristics, such as text descriptions or images.
This section analyzes whether it is reasonable to use complex geometries, including product spaces and overlapping spaces, in such a scenario.

For this purpose, we trained a classic DSSM model\footnote{We changed dense layers sizes in order to achieve the required embedding length and used more complex text tokenization with char bigrams, trigrams, and words, instead of just char trigrams.}~\cite{huang2013learning} on a private Wikipedia search dataset consisting of 491044 pairs (search query, relevant page), examples are given in Table~\ref{examples-dssm-table} in Appendix. All queries are divided into train, validation, and test sets, and for each signature, the optimal iteration was selected on the validation set. Table~\ref{dssm-table} compares all models for two embedding sizes. For short embeddings, we see that a product space based on spherical geometry is useful, and overlapping spaces have comparable quality. However, for large ``industrial size'' dimensions, the best results are achieved with the standard dot product, questioning the utility of complex geometries in the case of large dimensions. 

Note that in DSSM-like models, the most time-consuming task is model training. Hence, training multiple models for choosing the best configuration can be infeasible. Therefore, for small dimensions, overlapping spaces can be preferable over product spaces since they are universal and do not require parameter tuning. Moreover, calculating element embeddings is usually more time-consuming than calculating distances. Hence, even though calculating distances in the overlapping space has larger complexity than in simpler spaces, it does not have a noticeable effect in real applications.


\subsection{Synthetic bipartite graph reconstruction}\label{sec:synthetic}

Let us additionally illustrate that some graph structures are hard to embed into (considered) metric spaces. Our intuition is that the dot product is suitable for modelling \emph{popularity} of items since it can be naturally reflected using vector norms. Thus, we consider a situation when there are a few objects that are more popular than the other ones. To model this, we consider a synthetic bipartite graph with two sets of sizes 20 and 700 with 5\% edge probability (isolated nodes were removed, and the remaining graph is connected). Clearly, in the obtained graph, there are a few popular nodes and many nodes of small degrees. Figure~\ref{fig:bg} visualizes the obtained graph. Table~\ref{bg-table} compares the performance of the best metric space with the dot-product performance.   As we can see, this experiment confirms our intuition that some specific graphs are hard to embed into metric spaces, even with distortion loss. We also see that our OS-Mixed approach gives the best result with a margin. Thus, this experiment additionally confirms the universality of the proposed approach. We also observe that the optimization of WIPS is highly unstable on this dataset, see Table~\ref{bg-table-2} for details.


\begin{minipage}{0.5\textwidth}
        \centering
        \captionof{table}{Bipartite graph reconstruction}\label{bg-table}
        \begin{tabular}{l|lc}
            \toprule
            \textbf{} & \textbf{mAP} & \textbf{distortion} \\
            \midrule
            $E_{10}$
                & 0.777 & 0.094 \\
            $H_{10}$
                & 0.794 & 0.095 \\
            $S_{10}$
                & 0.796 & 0.096 \\
            $H_5^2$
                & 0.799 & 0.090 \\
            $S_5^2$
                & 0.796 & 0.094 \\
            $H_5 \times S_5$
                & 0.798 & 0.090 \\
            $H_2^5$
                & 0.761 & 0.086 \\
            $S_2^5$
                & 0.773 & 0.092 \\
            $H_2^2 \times E_2 \times S_2^2$
                & 0.796 & 0.089 \\
            \midrule
            $O_{l1}, t=0$
                & 0.824 & 0.094 \\
            $O_{l1}, t=1$
                & 0.803 & 0.082 \\
            $O_{l2}, t=1$
                & 0.814 & 0.092 \\
            \midrule
            \text{best metric space}
                & 0.824 & 0.082 \\
            $c-\text{dot}$
                & 0.863 & 0.079  \\
            \midrule 
            $c-\text{wips}$
                & 1 & 0.091  \\
            $O_{mix-l1}, t=1$
                & 0.986 & 0.083 \\
            $O_{mix-l2}, t=1$
                & 1 & 0.070 \\
            \bottomrule
        \end{tabular}
        
\end{minipage}\begin{minipage}{0.5\textwidth}
    \includegraphics[width=\textwidth]{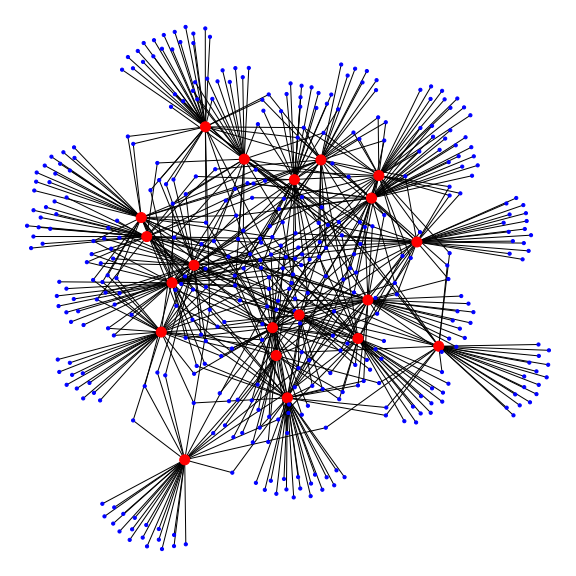}
    \captionof{figure}{Graph visualization. Red (big) nodes belong to the smaller part.}\label{fig:bg}
\end{minipage}

\begin{table}[t]
    \caption{WIPS distortion (5 restarts; best learning rate)}
    \label{bg-table-2}
    \centering
    \begin{small}
    \begin{tabular}{l|llll}
        \toprule
        \textbf{} & \textbf{avg.} & \textbf{worst} & \textbf{best} & \textbf{std} \\
        \midrule
        $c-\text{wips}$
            & 0.092 & 0.100 & 0.078 & 0.0078 \\
        $O_{mix-l2}, t=1$
            & 0.071 & 0.074 & 0.069 & 0.0018 \\
        \bottomrule
    \end{tabular}
    \end{small}
\end{table}

\section{Conclusion}\label{sec:conclusion}

This paper proposed the new concept of overlapping spaces that do not require signature brute-forcing and have better or comparable performance relative to the best product space in the graph reconstruction task. Improvements are observed for both global distortion and local mAP loss functions. 
An important advantage of our method is that it allows us to easily incorporate new distances or similarities as building blocks. The obtained overlapping-mixed non-metric measure achieves the best results for both distortion and mAP. We also evaluated the proposed overlapping spaces in the DSSM setup, and in the case of short embeddings, product space gives a better result than standard spaces, and the OS is comparable to it. In the case of long embeddings, no profit from complex spaces was found.

\begin{ack}

    The authors would like to thank Egor Samosvat for fruitful discussions.
    
    
    The work of Liudmila Prokhorenkova is partially supported by the Ministry of Education and Science
    of the Russian Federation in the framework of MegaGrant 075-15-2019-1926 and by the Russian
    President grant supporting leading scientific schools of the Russian Federation NSh2540.2020.1.
    Besides that, the funding was provided exclusively by the companies listed in the authors’ affiliation.

    
\end{ack}


\newpage

\bibliographystyle{ACM-Reference-Format}
\bibliography{arXiv.bib}


\newpage

\section*{Checklist}

\begin{enumerate}

\item For all authors...
\begin{enumerate}
  \item Do the main claims made in the abstract and introduction accurately reflect the paper's contributions and scope?
    \answerYes{}
  \item Did you describe the limitations of your work?
    \answerYes{See sections \ref{sec:experiments} and \ref{sec:conclusion}.}
  \item Did you discuss any potential negative societal impacts of your work?
    \answerNo{Any possible impact depends on the usage, not the approach itself.}
  \item Have you read the ethics review guidelines and ensured that your paper conforms to them?
    \answerYes{Yes, the paper conforms General Ethical Conduct: for example, we do not provide users' data from Section~\ref{dssm_exp}.}
\end{enumerate}

\item If you are including theoretical results...
\begin{enumerate}
  \item Did you state the full set of assumptions of all theoretical results?
    \answerYes{}
	\item Did you include complete proofs of all theoretical results?
    \answerYes{}
\end{enumerate}

\item If you ran experiments...
\begin{enumerate}
  \item Did you include the code, data, and instructions needed to reproduce the main experimental results (either in the supplemental material or as a URL)?
    \answerYes{The link to the code is provided and the description is given in Appendix.  Users' requests for the experiment in Section~\ref{dssm_exp} are not provided due to privacy rules; public graph datasets are sufficient to reproduce the main results.}
  \item Did you specify all the training details (e.g., data splits, hyperparameters, how they were chosen)?
    \answerYes{Both in Appendix and code.}
	\item Did you report error bars (e.g., with respect to the random seed after running experiments multiple times)?
    \answerNo{In the experiments with graph reconstruction, stability is ensured by using different datasets, in the experiments with DSSM, restarting does not change the final results in any significant way.}
	\item Did you include the total amount of compute and the type of resources used (e.g., type of GPUs, internal cluster, or cloud provider)?
    \answerYes{Technical requirements for our implementation are provided with code, small experiments can be reproduced on a regular computer.}
\end{enumerate}

\item If you are using existing assets (e.g., code, data, models) or curating/releasing new assets...
\begin{enumerate}
  \item If your work uses existing assets, did you cite the creators?
    \answerYes{See section \ref{graph_rec}.}
  \item Did you mention the license of the assets?
    \answerNo{No, although the data is available now, this may change for reasons beyond our control.}
  \item Did you include any new assets either in the supplemental material or as a URL?
    \answerYes{See the WLA6 dataset.}
  \item Did you discuss whether and how consent was obtained from people whose data you're using/curating?
    \answerYes{Indirectly, it is widely known that Wikipedia uses the Creative Commons license.}
  \item Did you discuss whether the data you are using/curating contains personally identifiable information or offensive content?
    \answerNo{We only collect the Wikipedia category graph.}
\end{enumerate}

\item If you used crowdsourcing or conducted research with human subjects...
\begin{enumerate}
  \item Did you include the full text of instructions given to participants and screenshots, if applicable?
    \answerNA{}
  \item Did you describe any potential participant risks, with links to Institutional Review Board (IRB) approvals, if applicable?
    \answerNA{}
  \item Did you include the estimated hourly wage paid to participants and the total amount spent on participant compensation?
    \answerNA{}
\end{enumerate}

\end{enumerate}

\appendix
\newpage

\section{Experimental setup}\label{sec:setup}

\subsection{Training details}

All models discussed in Section~\ref{graph_rec}
were trained with 2000 iterations. If more than one learning rate was used for a certain dataset (due to problems with the convergence of individual models), all the spaces were evaluated for all learning rates, and the best result was reported for each space. For distortion, the learning rate was 0.1 for all datasets except USCA312 (Cities), where we had 0.1 and 0.01. For mAP, the learning rate 0.1 was used for all datasets except USCA312 and CSPhDs, where we had 0.01 and 0.05 for both datasets.

For the experiments in Section~\ref{dssm_exp}
, we used 5000 iterations for short embeddings and 1000 for long ones (long embeddings converged faster). Hard-negative mining was not used for DSSM training.
Instead, large batches of 4096 random training examples (almost 1\% of the entire dataset) were used. During the learning process, only the training queries and documents were used. For evaluation, the nearest website was searched among all the documents. The training part was 90\% of the dataset, and the quality discrepancy between validation and test sets was quite small. Data samples are given in the table~\ref{examples-dssm-table}.

For the synthetic experiment in Section~\ref{sec:synthetic}
, for all spaces, the learning rates 0.1, 0.05, 0.01, 0.001 were used, and the best result was selected. We had 2000 and 1000 iterations for distortion and mAP, respectively. 


\subsection{WLA6 dataset details}\label{sec:WLA-dataset}

As described in the main text, this dataset is obtained by running the breadth-first search algorithm on the category graph
of the English-language Wikipedia (\url{https://en.wikipedia.org/wiki/Special:CategoryTree}), starting from the vertex (category) ``Linear algebra'' and limited to the depth 6 (Wikipedia Linear Algebra 6). We provide this graph along with the texts (names) of the vertices (categories). The resulting graph is very close to being a tree, although there are some cycles. Predictably, hyperbolic space gives a significant profit for this graph, while using product spaces gives almost no additional advantage. The purpose of using this dataset is to check our conclusions on data other than those used in~\cite{gu2018learning} and to evaluate overlapping spaces on a dataset where product spaces do not provide quality gains.

\section{Additional experimental results}

\subsection{Our implementation of product spaces vs original one}

Table~\ref{original-table} compares our implementation with the results reported in~\cite{gu2018learning}. It should be noted that we have significantly different algorithms with differing numbers of iterations. 

The optimal values of distortion obtained with our algorithm (except for the USCA312 dataset) are comparable and usually better than those reported in~\cite{gu2018learning}. On USCA312, the obtained distortion is orders of magnitude better, which can be caused by the proper choice of the learning rate (in our experiments on this dataset, this choice significantly affected the results).
These results indicate that our solution is a good starting point to compare different spaces and similarities. 

For mAP, we optimize the proxy-loss, in contrast to the canonical implementation, where both metrics were specified for models trained with distortion. Clearly, the results are more stable for our approach: we do not have such a large spread of values for different spaces. We noticed that directly optimizing ranking losses leads to significant improvements.

\begin{table*}
    \caption{Graph reconstruction: original product spaces vs our implementation}
    \label{original-table}
    \centering
    \begin{tabular}{l|cccccccc}
        \toprule
            & \multicolumn{2}{c}{\textbf{USCA312}} & \multicolumn{2}{c}{\textbf{CS PhDs}} & \multicolumn{2}{c}{\textbf{Power}} & \multicolumn{2}{c}{\textbf{Facebook}} \\
        \cmidrule(r){2-9} 
            & Canon. & Our & Canon. & Our & Canon. & Our & Canon. & Our \\     
        \midrule
        \multicolumn{9}{c}{Distortion} \\
        \midrule
        $E_{10}$
            & 0.0735 & 0.0032 &
                0.0543 & 0.0475 &
                    0.0917 & 0.0408 &
                        0.0653 & 0.0487 \\
        $H_{10}$
            & 0.0932 & 0.0111 &
                0.0502 & 0.0443 &
                    0.0388 & 0.0348 &
                        0.0596 & 0.0483 \\

        $S_{10}$
            & 0.0598 & 0.0095 &
                0.0569 & 0.0503 &
                    0.0500 & 0.0450 &
                        0.0661 & 0.0540 \\
        
        $H_5\times H_5$
            & 0.0756 & 0.0057 &
                0.0382 & 0.0345 &
                    0.0365 & 0.0255 &
                        0.0430 & 0.0372 \\
        
        $S_5 \times S_5$
            & 0.0593 & 0.0079 &
                0.0579 & 0.0492 &
                    0.0471 & 0.0433 &
                        0.0658 & 0.0511 \\
        
        $H_5 \times S_5$
            & 0.0622 & 0.0068 &
                0.0509 & 0.0337 &
                    0.0323 & 0.0249 &
                        0.0402 & 0.0318 \\

        $H_2^5$
            & 0.0687 & 0.0059 &
                0.0357 & 0.0344 &
                    0.0396 & 0.0273 &
                        0.0525 & 0.0439 \\
        
        $S_2^5$
            & 0.0638 & 0.0072 &
                0.0570 & 0.0460 &
                    0.0483 & 0.0418 &
                        0.0631 & 0.0489 \\
        
        $H_2^2 \times E_2 \times S_2^2$
            & 0.0765 & 0.0044 &
                0.0391 & 0.0345 &
                    0.0380 & 0.0299 &
                        0.0474 & 0.0406 \\
        
        \midrule
        \multicolumn{9}{c}{mAP} \\
        \midrule
        
        $E_{10}$
            & & 0.9290 &
                0.8691 & 0.9487 &
                    0.8860 & 0.9380 &
                        0.5801 & 0.7876 \\
        
        $H_{10}$
            & & 0.9173 &
                0.9310 & 0.9399 &
                    0.8442 & 0.9385 &
                        0.7824 & 0.7997 \\

        $S_{10},$
            & & 0.9254 &
                0.8329 & 0.9578 &
                    0.7952 & 0.9436 &
                        0.5562 & 0.7868 \\
        
        $H_5\times H_5$
            & & 0.9247 &
                0.9628 & 0.9481 &
                    0.8605 & 0.9415 &
                        0.7742 & 0.8084 \\

        $S_5 \times S_5$
            & & 0.9231 &
                0.7940 & 0.9662 &
                    0.8059 & 0.9466 &
                        0.5728 & 0.7891 \\
        
        $H_5 \times S_5$
            & & 0.9316 &
                0.9141 & 0.9654 &
                    0.8850 & 0.9467 &
                        0.7414 & 0.8087 \\
        
        $H_2^5$
            & & 0.9364 &
                0.9694 & 0.9671 &
                    0.8739 & 0.9508 &
                        0.7519 & 0.7979 \\
        
        $S_2^5$
            & & 0.9281 &
                0.8334 & 0.9714 &
                    0.8818 & 0.9521 &
                        0.5808 & 0.7915 \\

        $H_2^2 \times E_2 \times S_2^2$
            & & 0.9391 &
                0.8672 & 0.9611 &
                    0.8152 & 0.9486 &
                        0.5951 & 0.7970 \\
        
        \bottomrule
    \end{tabular}
\end{table*}

\subsection{Parametrization of spherical space}\label{sec:sphere-param}

In Tables~\ref{ditortion-table}
and~\ref{map-table}
of the main text, we used hyperspherical parameterization of spherical subspaces in product spaces since we fixed the number of stored values for each space. Here, in Tables~\ref{spheres-ditortion-table} and ~\ref{spheres-map-table}, we present the extended results, where we fix the mathematical dimension of product spaces and use $d+1$ parameters and simple mappings from Section~\ref{sec:overlapping}, equation~\eqref{eq:mapping},

as done in~\cite{gu2018learning}.
We can see that our implementation gives results comparable to the original ones in distortion setup and significantly better for mAP, which is associated with using the proxy-loss instead of distortion.

\begin{table}[h]
    \caption{Graph reconstruction with distortion loss, top results are highlighted, metrics only}
    \label{spheres-ditortion-table}
    \centering
    \begin{tabular}{l|ccccc}
        \toprule
        \textbf{Signature}  & \textbf{USCA312} & \textbf{CS PhDs} & \textbf{Power} & \textbf{Facebook} & \textbf{WLA6} \\
        \midrule
        $E_{10}$
            & \red{$\textbf{0.00318}$} & 0.0475 & 0.0408 & 0.0487 & 0.0530 \\
        $H_{10}$
            & 0.01114 & 0.0443 & 0.0348 & 0.0483 & \textbf{0.0279} \\
        $S_{10}$
            & 0.00951 & 0.0503 & 0.0450 & 0.0540 & 0.0589 \\
        $H_5^2 \equiv H_5\times H_5 $
            & 0.00573 & 0.0345 & 0.0255 & 0.0372 & \textbf{0.0279} \\
        $S_5 \times S_5 \equiv S_5^2$
            & 0.00792 & 0.0492 & 0.0433 & 0.0511 & 0.0585 \\
        $H_5 \times S_5$
            & 0.00681 & \textbf{0.0337} & \textbf{0.0249} & \textbf{\red{0.0318}} & 0.0296 \\
        $H_2^5$
            & 0.00592 & 0.0344 & 0.0273 & 0.0439 & 0.0356 \\
        $S_2^5$
             & 0.00720 & 0.0460 & 0.0418 & 0.0489 & 0.0549 \\
        $H_2^2 \times E_2 \times S_2^2$
            & 0.00436 & 0.0345 & 0.0299 & 0.0406 & 0.0405 \\
        $O_{l1}, t=0$
            & \textbf{0.00356} & 0.0368 & 0.0281 & 0.0458 & 0.0286 \\
        $O_{l1}, t=1$
            & \blue{$\textbf{0.00330}$} & \textbf{\red{0.0300}} & \textbf{\red{0.0231}} & \textbf{0.0371} & \textbf{\red{0.0272}} \\
        $O_{l2}, t=1$
            & 0.00530 & \textbf{\blue{0.0328}} & \textbf{\blue{0.0246}} & \textbf{\blue{0.0324}} & \textbf{\blue{0.0278}} \\
        \bottomrule
    \end{tabular}
\end{table} 

\begin{table}[H]
    \caption{Graph reconstruction with mAP ranking loss, top results are highlighted, metrics only}
    \label{spheres-map-table}
    \centering
    \begin{tabular}{l|ccccc}
        \toprule
        \textbf{Signature} & \textbf{USCA312} & \textbf{CS PhDs} & \textbf{Power}  & \textbf{Facebook} & \textbf{WLA6}        \\ \midrule
        $E_{10}$
            & 0.9290 & 0.9487 & 0.9380  & 0.7876 & 0.7199 \\
        $H_{10}$
            & 0.9173 & 0.9399 & 0.9385 & 0.7997 & 0.9617 \\
        $S_{10}$
            & 0.9254 & 0.9578 & 0.9436 & 0.7868 & 0.7287  \\
        $H_5^2$
            & 0.9247 & 0.9481 & 0.9415 & 0.8084 & \textbf{0.9682} \\
        $S_5^2$
            & 0.9231 & 0.9662 & 0.9466 & 0.7891 & 0.7353  \\
        $H_5 \times S_5$
            & 0.9316 & 0.9654 & 0.9467 & 0.8087 & \textbf{\red{0.9779}} \\
        $H_2^5$
            & 0.9364 & 0.9671 & 0.9508 & 0.7979 & 0.8597 \\
        $S_2^5$
            & 0.9281 & 0.9714 & 0.9521 & 0.7915 & 0.7346 \\
        $H_2^2 \times E_2 \times S_2^2$
            & 0.9391 & 0.9611 & 0.9486 & 0.7970  & 0.6796 \\
        $O_{l1}, t=0$
            & \textbf{\red{0.9522}} & \textbf{\blue{0.9879}} & \textbf{0.9728} & \textbf{0.8093} & 0.6759 \\
        $O_{l1}, t=1$
            & \textbf{\red{0.9522}} & \textbf{0.9904} & \textbf{\blue{0.9762}} & \textbf{\blue{0.8185}} & 0.9598 \\
        $O_{l2}, t=1$
            & \textbf{\red{0.9522}} & \textbf{\red{0.9938}} & \textbf{\red{0.9907}} & \textbf{\red{0.8326}} & \textbf{\blue{0.9694}} \\
        \bottomrule
    \end{tabular}
\end{table}

\subsection{Other ways of converting distances to probabilities}

For the proxy-loss, we additionally experimented with other ways of converting distances to probabilities. 
Let us write $L_{proxy}$ in the general form:

\begin{equation}\label{proxyloss-general}
\begin{aligned}
    L_{proxy} &= - \sum\limits_{(v, u) \in E} \log \mathrm{P}((v, u) \in E) = - \sum\limits_{(v, u) \in E} \log \frac{t\big{(}d_U(f(v), f(u)))\big{)}}{\sum\limits_{w \in V}t\big{(}d_U(f(v), f(w))\big{)}}\,,
\end{aligned}
\end{equation}
where $t(d)$ is a function that decreases with distance $d$. We compare the following alternatives for $t(d)$:
\begin{align*}
    t_1(d) = \exp(-d),
    t_2(d) = \exp\left(\frac{1}{\min(d, d_0)}\right),
    t_3(d) = \frac{1}{\min(d, d_0)}\,,
\end{align*}
where $d_0$ is a small constant.

Recall that $t_1$ was used in the main text and it seems to be the most natural choice.\footnote{Note that this is the softmax over the inverted distances.} Table~\ref{sdplc-table} compares the options and shows that the best results are indeed achieved with $t_1$. 

\begin{table}
\vspace{10pt}
  \caption{Comparison of proxy-losses, mAP}
  \label{sdplc-table}
  \centering
  \begin{tabular}{l|llllll}
    \toprule
     & \multicolumn{3}{c}{\textbf{USCA312}} & \multicolumn{3}{c}{\textbf{CS PhD}} \\
    \cmidrule(r){2-4}  \cmidrule(r){5-7}
    \hspace{50pt}$P \sim$
        & $e^{-d}$ & $e^{1/d}$ & $1/d$
            & $e^{-d}$ & $e^{1/d}$ & $1/d$ \\
    \midrule
    $E_{10}$
        & 0.929 & 0.911 & 0.899
            & 0.949 & 0.956 & 0.831 \\
    $H_{10}$
        & 0.917 & 0.807 & 0.885
            & 0.940 & 0.749 & 0.764 \\
    $S_{10}$
        & 0.925 & 0.797 & 0.838
            & 0.958 & 0.572 & 0.689 \\
    $H_5^2$
        & 0.925 & 0.890 & 0.883
            & 0.948 & 0.976 & 0.723 \\
    $S_5^2$
        & 0.923 & 0.802 & 0.858
            & 0.966 & 0.748 & 0.775 \\
    $H_5 \times S_5$
        & 0.932 & 0.838 & 0.865
            & 0.965 & 0.804 & 0.721 \\
    $H_2^5$
        & 0.936 & 0.896 & 0.903
            & 0.967 & 0.998 & 0.823 \\
    $S_2^5$
        & 0.928 & 0.856 & 0.871
            & 0.971 & 0.876 & 0.881 \\
    $H_2^2 \times E_2 \times S_2^2$
        & 0.939 & 0.872 & 0.865
            & 0.961 & 0.884 & 0.689 \\
    \midrule
    $O_{l1}, t=0$
        & 0.952 & 0.933 & 0.872
            & 0.988 & 0.961 & 0.762 \\
    $O_{l1}, t=1$
        & 0.952 & 0.947 & 0.877
            & 0.990 & 0.963 & 0.815 \\
    $O_{l2}, t=1$
        & 0.952 & 0.939 & 0.880
            & 0.994 & 0.979 & 0.810 \\
    \midrule
    $c-\text{dot}$
        & 1     & 1     & 0.777
            & 1     & 0.999 & 0.917 \\
    \bottomrule
  \end{tabular}
\end{table}

\begin{table}[t]
    \caption{Search query examples}
    \label{examples-dssm-table}
    \centering
        \begin{tabular}{ll}
        \toprule
        \textbf{Query} & \textbf{Web site} \\
        \midrule
        Kris Wallace
            & en.wikipedia.org/wiki/Chris\_Wallace \\
        1980: Mitsubishi produces one million cars...//
            & en.wikipedia.org/wiki/Mitsubishi\_Motors \\
        code napoleon
            & en.wikipedia.org/wiki/Napoleonic\_Code \\
        \bottomrule
    \end{tabular}
\end{table}

\subsection{Analysis of depth in overlapping spaces}\label{sec:os-more-t}

Distortion graph reconstruction results for all possible $t \leq \log_2(d) = \log_2(10) \sim 3.3$ are provided in Table~\ref{depth-ditortion-table} for completeness. The results below confirm our hypothesis that the reconstruction distortion improves with increasing $t$.

\begin{table}
    \caption{Distortion graph reconstruction for different overlapping spaces}
    \label{depth-ditortion-table}
    \centering
    \begin{small}   
    \begin{tabular}{l|cccccc}
        \toprule
        \textbf{Signature}  & \textbf{USCA312} & \textbf{CS PhDs} & \textbf{Power} & \textbf{Facebook} & \textbf{WLA6} & \textbf{EuCore} \\
        \midrule
        
        $O_{l1}, t=0$
            & 0.00324 & 0.0368 & 0.0281 & 0.0458 & 0.0286 & 0.1141 \\
        \midrule
        $O_{l1}, t=1$
            & 0.00325 & 0.0300 & 0.0231 & 0.0371 & 0.0272 & 0.1117 \\
        $O_{l1}, t=2$
            & 0.00296 & 0.0335 & 0.0262 & 0.0309 & 0.0273 & 0.1114 \\
        $O_{l1}, t=3$
            & 0.00257 & 0.0273 & 0.0209 & 0.0313 & 0.0246 & 0.1098 \\
        \midrule
        $O_{l2}, t=1$
            & 0.00530 & 0.0328 & 0.0246 & 0.0324 & 0.0278 & 0.1127 \\
        $O_{l2}, t=2$
            & 0.00596 & 0.0303 & 0.0256 & 0.0312 & 0.0278 & 0.1117 \\
        $O_{l2}, t=3$
            & 0.00303 & 0.0343 & 0.0240 & 0.0302 & 0.0279 & 0.1119 \\
        \bottomrule
    \end{tabular}
    \end{small}
\end{table}

\subsection{Analysis of learned weights}\label{sec:os-weights}

While analyzing the trained weights we have made several observations:
    
\begin{enumerate}
    \item We see that OS does not learn a pure product space. In particular, on the CS PhDs dataset we get $${d_{O_{l=1}, t=0} \propto 0.37 d_H + 0.63 d_S},$$ which is significantly better than both $d_S$ and $d_H$ separately.
    
    \item If for $t=0$ there is a space with a noticeably larger weight compared to the other ones, then the space of same type often makes the largest contribution for $t = 1$ too. For example, in USCA312, $$d_{O_{l1}, t=0} \propto \textbf{0.90} d_E + 0.05 d_H + 0.05 d_S,$$ and the weights of the Euclidean subdistances for $d_{O_{l1}, t=1}$ (normalized, $\sum w_i = 1$) are 0.6, 0.15, 0.1.
    \item However, a space that is absent for $t = 0$ can appear for $t = 1$. For example, in the Power dataset, $${d_{O_{l1}, t=0} \propto 0.37 d_H + 0.63 d_S},$$ $${d_{O_{l1}, t=1} \propto \textbf{0.1} d_E(l_1^0, r_1^0) + 0.5 d_R(l_1^0, r_1^0) + 0.4 d_H(l_1^1, r_1^1)},$$ 
            where $l_1^0 = l[0..5], l_1^1 = l[6..10]$. 

            \item Finally, we noticed that almost always, more than half of the weights are near-zero, which allows one to remove unnecessary distances and improve efficiency.
            
        \end{enumerate}
        
\section{Proof of Statement~\ref{st:metric}}\label{sec:os-is-metric-proof}

To prove that $d(x,y)$ is a metric distance, we need to show that it is symmetric, nonnegative, equals zero only when $x=y$, and satisfies the triangle inequality.

Consider an overlapping space: 
$$d_{O}(l, r) = \text{Agg}(d_{D_1}(\cdot,\cdot), \ldots, d_{D_k}(\cdot,\cdot)),$$
where $\text{Agg}$ is $l1$ or $l2$ aggregation and $d_{D_i}$ are base distances applied to subsets of coordinates.

Symmetry of $d_O$ follows from symmetry of base distances $d_{D_i}$. Obviously, we have $d_O(x,y) \ge 0$ and $d_O(x,x) = 0$. The inequality $d_O(x,y) > 0$ for $x \neq y$ follows from the fact hat we use specific non-trival mappings $M_{D_i}$ and assume that together subsets of coordinates $p_i$ cover all coordinates (i.e., $\cup_{i=1}^k p_i = \{1, \ldots, d\}$).

Obviously, $l1$ aggregation (sum) preserves the triangle inequality. So, it remains to show this for $l2$. Assume that $d_1$ and $d_2$ satisfy the triangle inequality, nonnegative and let $d_{l2} = \sqrt{d_1^2 + d_2^2}$.

Let $c_1 := d_1(x,y)$, $c_2 := d_2(x,y)$, $a_1 = d_1(x,z)$, $a_2 = d_2(x,z)$, $b_1 = d_1(z,y)$, $b_2 = d_2(z,y)$.

We know that $c_1 \le a_1 + b_1$ and $c_2 \le a_2 + b_2$. Therefore,
\begin{equation}\label{eq:1}
c_1^2 + c_2^2 \le a_1^2 + a_2^2 + b_1^2 + b_2^2 + 2 a_1 b_1 + 2 a_2 b_2\,.
\end{equation}

We need to show
\[
\sqrt{c_1^2 + c_2^2} \le  \sqrt{a_1^2 + a_2^2} + \sqrt{b_1^2 + b_2^2}\,,
\]
\[
c_1^2 + c_2^2 \le a_1^2 + a_2^2 + b_1^2 + b_2^2 + 2\sqrt{a_1^2 + a_2^2}\sqrt{b_1^2 + b_2^2}\,.
\]
Taking into account Equation~\ref{eq:1}, it is sufficient to show
\[
2 a_1 b_1 + 2 a_2 b_2 \le 2\sqrt{a_1^2 + a_2^2}\sqrt{b_1^2 + b_2^2}\,,
\]
\[
a_1^2 b_1^2 + a_2^2 b_2^2 + 2 a_1 b_1 a_2 b_2 \le (a_1^2 + a_2^2)(b_1^2 + b_2^2)\,,
\]
\[
2 a_1 b_1 a_2 b_2 \le a_1^2 b_2^2 + a_2^2b_1^2\,,
\]
which is true.

Finally, note that that for three base distances we have $d_{l2} = \sqrt{ (\sqrt{d_1^2 + d_2^2})^2 + d_3^2} = \sqrt{d_1^2 + d_2^2 + d_3^2} $ and so we have proved the statement for an arbitrary number of terms.

\section{Additional illustrations for overlapping spaces}\label{sec:os-with-pictures}

\begin{figure}
    \centering

    \subfigure[Computing Euclidean distance]{\includegraphics[width=0.75\textwidth]{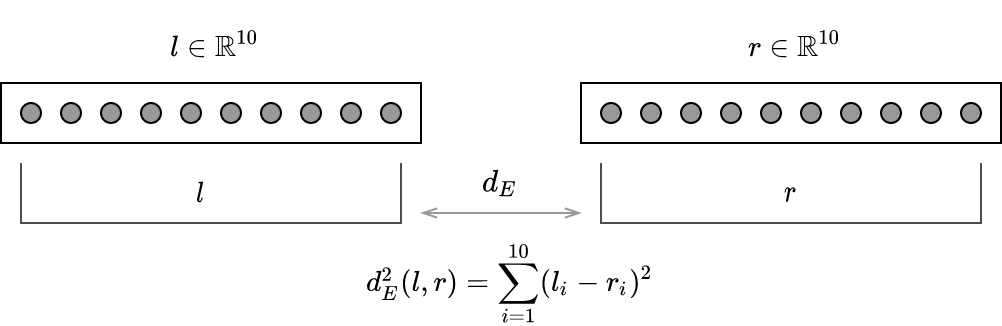}\label{fig:a}}

    \subfigure[Computing hyperbolic distance]{\includegraphics[width=0.75\textwidth]{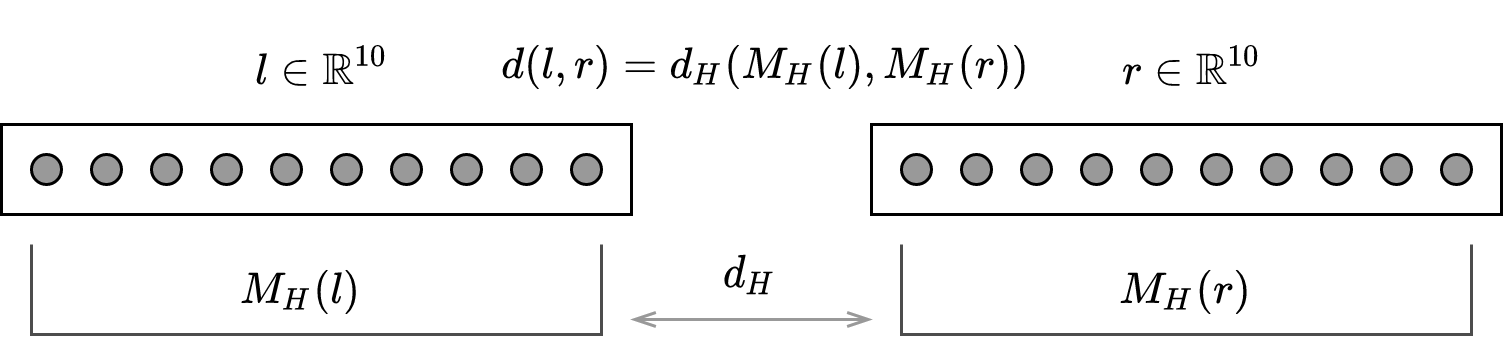}\label{fig:b}}
    
    \subfigure[Computing product space distance]{\includegraphics[width=0.75\textwidth]{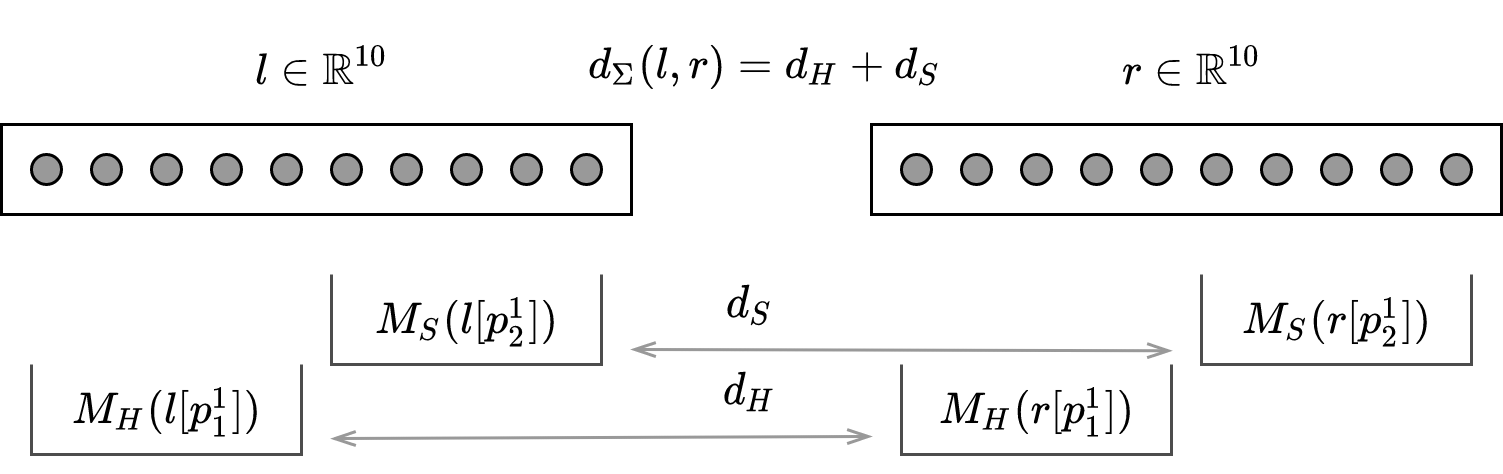}\label{fig:c}}

    \subfigure[Example of overlapping space distance]{
    \includegraphics[width=0.75\textwidth]{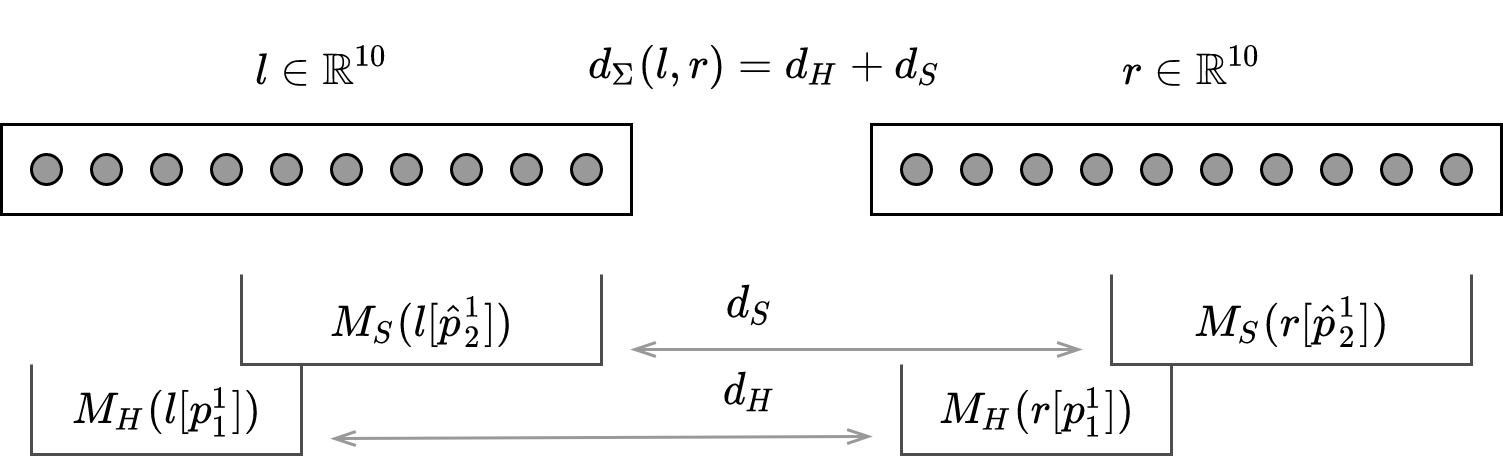} \label{fig:d}
    }
    \caption{Illustrating overlapping space with $d=10$ and $l1$ (sum) aggregation}
    \label{fig:simple_illustration}
\end{figure}

Figure~\ref{fig:simple_illustration} additionally illustrates the idea behind overlapping spaces. Namely, Figure~\ref{fig:a} shows standard Euclidean distance evaluation between two vectors $l$ and $r$. As shown in Figure~\ref{fig:b}, we add a differentiable mapping $M_H : \mathbb{R}^{10} \rightarrow H_{10}$ to calculate the distance in the hyperbolic space (we may do the same for the spherical space). Applying several mappings to different parts of $l$ and $r$, we may get any product space as shown in Figure~\ref{fig:c}. The last step is to allow the subsets of coordinates to \textit{overlap}, as shown in Figure~\ref{fig:d}, where the fifth coordinate is used simultaneously in two mappings. All such spaces with all possible intersections and base distances are called overlapping spaces.

\end{document}